\documentclass[pdflatex,default]{sn-jnl}

\jyear{2023}%

\theoremstyle{thmstyleone}%
\theoremstyle{thmstyletwo}%

\theoremstyle{thmstylethree}%

\usepackage{amssymb}
\usepackage{multicol}
\usepackage{lipsum}
\usepackage{mwe}
\usepackage{multirow}
\usepackage{graphicx}
\usepackage{caption}
\usepackage{arabtex}
\usepackage{utf8}
\setcode{utf8}

\begin{document}

\title[Poem Meter Classification of Recited Arabic Poetry]{Poem Meter Classification of Recited Arabic Poetry: Integrating High-Resource Systems for a Low-Resource Task}


\author[1]{\fnm{Maged S.} \sur{Al-Shaibani}}\email{g201381710@kfupm.edu.sa}

\author[1]{\fnm{Zaid} \sur{Alyafeai}}\email{g201080740@kfupm.edu.sa}

\author*[1,2]{\fnm{Irfan} \sur{Ahmad}}\email{irfan.ahmad@kfupm.edu.sa}


\author[3]{\fnm{Abdul Kareem Saleh} \sur{Al-Zahrani}}\email{akareem@kfupm.edu.sa}

\affil[1]{\orgdiv{Department of Computer Science}, \orgname{King Fahd University of Petroleum and Minerals}, \orgaddress{\street{Dhahran}, \postcode{31261}, \country{Saudi Arabia}}}
\affil[2]{\orgdiv{SDAIA--KFUPM Joint Research Center for AI}, \orgname{KFUPM}, \orgaddress{\street{Dhahran}, \postcode{31261}, \country{Saudi Arabia}}}
\affil[3]{\orgdiv{Islamic \& Arabic Studies Department}, \orgname{King Fahd University of Petroleum and Minerals}, \orgaddress{\street{Dhahran}, \postcode{31261}, \country{Saudi Arabia}}}




\abstract{Arabic poetry is an essential and integral part of Arabic language and culture. It has been used by the Arabs to spot lights on their major events such as depicting brutal battles and conflicts. They also used it, as in many other languages, for various purposes such as romance, pride, lamentation, etc. Arabic poetry has received major attention from linguistics over the decades. One of the main characteristics of Arabic poetry is its special rhythmic structure as opposed to prose. This structure is referred to as a meter. Meters, along with other poetic characteristics, are intensively studied in an Arabic linguistic field called "\textit{Aroud}". Identifying these meters for a verse is a lengthy and complicated process. It also requires technical knowledge in \textit{Aruod}. For recited poetry, it adds an extra layer of processing. Developing systems for automatic identification of poem meters for recited poems need large amounts of labelled data. In this study, we propose a state-of-the-art framework to identify the poem meters of recited Arabic poetry, where we integrate two separate high-resource systems to perform the low-resource task. To ensure generalization of our proposed architecture, we publish a benchmark for this task for future research.} 




\maketitle

\section{Introduction}
\label{introduction}

The last decade witnessed an unprecedented surge of advances in deep learning. The proposed deep architectures significantly advanced the state-of-the-art results in many tasks such as image recognition, image classification, and language modeling. This is due to the abundance of large data and computational machinery that was not previously available.

Speech recognition refers to the task of predicting spoken words from audio signals. The rapid developments occurring in this field feature a new trend of human communication with machines via virtual assistants. It has extensive uses due to its ease of communication and accessibility. It can be used in smart homes by interacting with voice-enabled devices to control house appliances. Customers can interact with voice-enabled software that mimics human interactions, etc. This new fashion of communication starts to reshape people's interactions with technology.
    
The early steps in the field of speech recognition used some rule-based and template-based methods. Later advances, then, tried to predict the words using Hidden Markov Models (HMMs) and statistical models. This approach was outperformed by neural networks producing better results. The rapid progress in neural networks introduced deep networks, which achieved state-of-the-art results that outperformed previous methods \citep{abdel2012applying,abdel2014convolutional}.
    
The advances in this area are still not fully explored for Arabic. Compared to other languages such as English, Arabic is on its initial steps \citep{al2018literature}. One of the possible reasons for such slow progress might be the scarcity of large-scale datasets. The first appearance of a large-scale acoustic dataset was the MGB-3 challenge proposed by \cite{ali2016mgb}. This challenge was also reiterated in 2017 \citep{ali2017speech} and 2019 \citep{ali2019mgb}. 

Arabic poetry, as in many other languages, is an integral and essential artifact of the language and its cultural heritage. Due to its importance, a dedicated field of study in Arabic literary, called \textit{Arud}, is advocated for poetry. It dates back to the rise of the Arabic and Islamic golden age. Such effort continued to the recent literature. Examples of such prominent work on the linguistic analysis of Arabic poetry include \cite{ziyovuddinova2021arud,manna2021metrics,paoli2001meters}.

Arabic poetry can be categorized into two main categories: modern and classical. Classical poetry is sometimes referred to as standard poetry, as it should satisfy strict poetic constraints. For instance, all the poem verses in a classical poetry should follow the same meter. There are a total of 16 meters in Arabic poetry. This constraint is relaxed in modern Arabic poetry. However, there should be a consistent sense of rhyme within the poem.

Compared to the Arabic prose processing, Arabic poetry processing has received less attention. However, recent efforts were directed toward approaching poetry processing tasks. For instance, \citet{al2020meter} and \citet{abandah2020classifying} proposed models to classify Arabic poem meters from textual sources. Other types of analysis study other topics, such as authorship attribution \cite{ahmed2015authorship} and emotion classification \cite{alsharif2013emotion}. 

Most of the research in this field addresses the written form of poetry, and processing poetry from other sources, such as poem recitation, is very limited. \citet{al2020automatic} investigated deep learning models for poem-meter classification from acoustic sources. He proposed an LSTM-based meter-classification model to classify poetry from sound signals. His proposed model classifies only 3 meters out of the 16 poem meters in Arabic. The model has been trained on a small dataset of 230 verses.

This research aims to advance Arabic poetry research by studying the task of meter classification from acoustic sources. Two approaches have been investigated. The first approach is an end-to-end system where the sound signals from recited verses are classified into meters. As it is a low-resource task with few labelled samples of recited poetry publicly available, we investigated a second approach where two systems trained on separate high-resource tasks are integrated with fine-tuning to perform poem-meter classification from recited poem verses. The contribution of this paper can be summarized as follows:

\begin{itemize}
    \item We present an end-to-end system to classify Arabic poem meters from recited poetry.
    \item We present a framework combining two separate high-resource systems to classify Arabic poem meters from recited poetry, which is considered a low-resource task.
    \item We present an evaluation dataset collected to serve as a benchmark for future research in this field. The presented systems were evaluated on the dataset. 
\end{itemize}

The rest of the paper is organized as follows: Section \ref{related_work} presents the related work. Section \ref{dataset} presents the dataset used in this research. Section \ref{methodology} illustrates the methodology followed to conduct this research. Section \ref{results} presents the results and analysis on the results. Finally, section \ref{conclusion} concludes the paper.

\section{Related Work}
\label{related_work}

To provide a proper background on the literature related to this research, this section surveyed the progress on poetry meter classification as well as speech recognition as both of these fields are investigated in this study.

\subsection{Automatic Speech Recognition}

The early developments related to automatic speech recognition (ASR) included a combination of Hidden Markov Models with Gaussian Mixed Models (HMMs-GMMs) after multiple rule-based attempts. The Sphinx system \citep{lee1988large} adopted this approach, achieving substantial progress compared to previous rule-based approaches. The Kaldi toolkit \citep{povey2011kaldi} was another attempt with more technical methods and feature extraction techniques. Other previous methods include template-based methods \cite{rabiner1979speaker}, knowledge-based methods \cite{de1986plan,kuhn1992corrections}, and dynamic time warping \cite{zhao2010speech} techniques. A review by \citet{radha2012review} provides a more detailed and a broader overview of the field up to that point on its development timeline. 
    
Due to the remarkable improvement leveraged by neural networks in various domains, \citet{dahl2011context} from Microsoft Research Institute introduced a deep neural network incorporated with a hidden Markov model to approach ASR. It achieved a noticeable performance gain as compared to the traditional methods. \citet{hannun2014deep} and their subsequent version \cite{amodei2016deep} employed a deeper network trained on large amounts of data. Their model is an end-to-end system based on Recurrent Neural Networks (RNNs). This model was trained on 5000 hours of speech recorded by 9600 speakers. This amount is overwhelming for languages or domains with limited datasets. This is not to mention the laborious effort required for labeling, time consumed, and being error prone.

The existing works in the literature of Arabic ASR extensively utilized knowledge-based methods to approach this task. Limited research has addressed Arabic ASR in an end-to-end manner. Probably the first research using an end-to-end framework for Arabic ASR was proposed by \citet{ahmed2019end}. Their model used bidirectional RNNs employing CTC as an objective function. Another recent work applying an end-to-end framework was proposed by \citet{alsayadi2021arabic}. Their model was the first end-to-end deep learning system for Arabic ASR producing diacritized texts. They employed a CNN-LSTM network with attention. They showed that their model outperformed traditional methods in this task. The review proposed by \citet{abdelhamid2020end} and \citet{dhouib2022arabic} chart the progress in this field for Arabic.

Since Arabic ASR started to receive attention, larger datasets dedicated to this task started to emerge. Examples include the MGB-3 challenge releases with three large scale datasets \citep{ali2016mgb,ali2017speech,ali2019mgb}. QASR dataset \citep{mubarak2021qasr} is also another massive dataset for dialectical Arabic. \citet{halabi2016modern} published an Arabic phonetic dataset for speech synthesis task.

\subsection{Self-Supervised and Transfer Learning}

To address issues related to the need of large datasets, researchers in the field of deep learning investigated alternatives including self-supervised learning \citep{liu2021self}  and transfer learning \citep{alyafeai2020survey,pan2009survey}. Self supervised learning refers to the process of training a large model on a large amount of unlabeled data \citep{liu2021self}. These models can then be fine-tuned on downstream tasks in a process referred to as transfer learning. Transfer learning, in general, is the process of moving knowledge from a trained model to another model. This can happen in many aspects such as fine-tuning, zero or few shot learning \citep{alyafeai2020survey}. Both of these approaches has been recently employed in many tasks in the field of Natural Language Processing (NLP), such as language modeling and text classification achieving state-of-the-art results in various tasks. Examples of these models are BERT \citep{devlin2018bert}, GPT-2 \citep{radford2019language}, and its successor GPT-3 \citep{brown2020language}. The prominent models following this methodology in the field of speech recognition is Wav2Vec \citep{schneider2019wav2vec} and Wav2Vec2.0 \citep{baevski2020wav2vec}.

Investigating and studying Wav2Vec2.0 as a state-of-the-art model for speech recognition, \citet{yi2020applying} compared its performance on many languages, including Arabic where the authors used a telephone conversation dataset of 8000 samples. Furthermore, \citet{bakheet2021improving} trained Wav2Vec from scratch on the Arabic Mozilla Common Voice dataset and reported a noticeable improvement on the accuracy and reduction in word error rates compared to previous approaches reported in the literature.

Employing these recent advances in Arabic speech recognition research seems to be very limited specially for automatic speech recognition task. \citet{nassif2019speech} conducted a broad systematic literature review for automatic speech recognition systems for various languages and Arabic is not reported within these languages. \citet{elnagar2021systematic} surveyed the literature for dialectial Arabic identification and detection from speech. They reported multiple studies conducted on speech recognition, but only a few of them utilized deep learning. Additionally, the use of transfer learning and fine-tuning was not reported. Other reviews \citep{abdelhamid2020end,algihab2019arabic,al2018literature} also provide a few studies on Arabic speech recognition systems using deep networks. This shows that this field is lacking for Arabic.

\subsection{Meter Classification}

The task of poetry meter classification has received attention in many languages due to the cultural importance of poetry. \citet{haider2021metrical} presented a poetry corpus for English and German poetry with machine learning baselines. \citet{yousef2019learning} presented an LSTM-based approach to identify English poem meters, as well as Arabic, from textual sources. They also introduced a large dataset for this task. \citet{agirrezabal2016machine} studied the metrical structures of acoustic English poetry using various NLP methods. \citet{rajan2022pomet} proposed POMET, a recorded set of Malayalam poetry consisting of 8 meters. They also introduced baseline experiments on this dataset. \citet{rajan2018poetic} and \citet{rajan2020poetic} proposed neural network-based methods along with musical features to approach Malayalam acoustic poetry meter classification. \citet{hamidi2009automatic} presented a classification system for acoustic Persian poetry utilizing SVMs. 

For Arabic, an extensive survey on poem-meter classification is introduced in the following paragraphs. The work on this task can be divided into two major parts based on the input source: acoustic and textual. Most of the existing research, however, studied meter classification from textual sources. Little work has addressed meter classification from audio signals of Arabic poems. This can be partly attributed to the lack of large acoustic datasets. 

\citet{al2020meter} proposed a GRU-based model to classify textual poetry to its meters. The MetRec dataset \cite{al2020metrec} consisting of 55,400 verses from the most well-known 14 meters in Arabic poetry was used for experimentation. 
Their approach does not require diacritizing the text, which was a preliminary step to approach this task in previous research. They reported an accuracy of more than 94\%.

\citet{yousef2019learning} proposed an LSTM-based model to classify Arabic and English poetry meters. In their work on Arabic, they investigated various configurations regarding diacritics and other parameters. They used a large dataset of more than 1.7 million verses collected from large Arabic poetry repositories from the web. In contrast to MetRec, their dataset contains all the 16 well-defined Arabic poem meters. The authors reported accuracy of around 96\%.

\citet{abandah2020classifying} proposed an LSTM-based approach to classify Arabic poetry. They adapted the dataset developed by \citet{yousef2019learning} to add samples from the 'prose' class. They developed a model that is capable of classifying a given text into 17 classes, 16 of which are the well-defined meters, while the seventeenth class is prose. In their model, the use of diacritics is optional. The results they reported are close to 95\% for F1-score.

\citet{saleh2012arabic} filed a patent for a system that automatically identifies the poem meters from either acoustic or textual sources. The model developed to classify meters from acoustic data is transcription-based, where the transcription model used is a composition of Hidden Markov Models (HMM) and Mel-frequency Cepstral Coefficients (MFCCs) features followed by a neural network classification layer. There are no further evaluation details available in the patent description.

\citet{al2020automatic} compared LSTMs and SVMs performance to classify poetry from acoustic sources. They experimented with multiple configurations of different features extracted from the sound signal. Their evaluation was conducted on a small dataset of 230 verses that is prepared with two human subjects. Verses in this dataset belong to only 3 meters out of the well-defined 16 meters. They showed that SVMs perform better with MFCC on their dataset.
        
\section{Datasets}
\label{dataset}

Two datasets have been employed in this research.
The first dataset is a manually collected dataset to perform the experimentation and evaluate the techniques. Its main purpose is to train, fine tune and evaluate the proposed models. It has been developed using special instruments that produce high-quality audio. It has been further processed with special software to generate better audio signals. Throughout this paper, we will refer to this dataset as the baseline dataset.

The second set, on the other hand, is collected mainly from the YouTube platform. The purpose of this set is to test the generalization of the proposed models trained on the baseline dataset. This set is publicly available and could be used as a benchmark for this task. Throughout this research, we will refer to this public set as the benchmark evaluation set.

The following two subsections provide an overview of these datasets.

\subsection{The Baseline Dataset}

This dataset consists of 3,668 transcribed acoustic files. Each file holds only one \textit{Bait} (a \textit{Bait} is a poem verse in English) that belongs to one meter. The dataset is 9 hours long making it, to the best of our knowledge, the largest dataset for Arabic poetry. Table \ref{table:training_dataset_statistics} illustrates some useful statistics about this dataset.

\begin{figure}
    \centering
    \includegraphics[width=12cm,height=12cm,keepaspectratio]{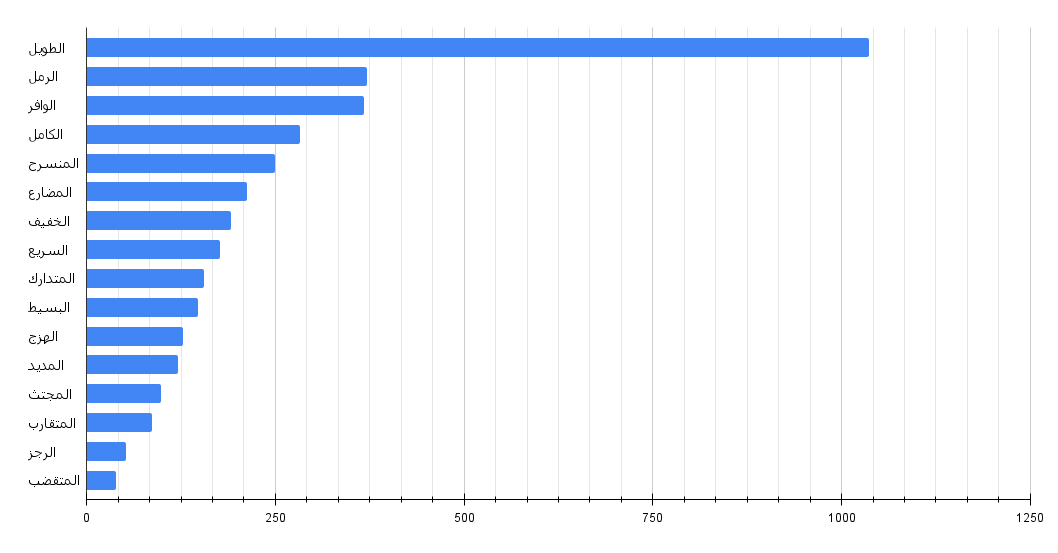}
    \caption{Meter distribution for the 16 poem meters in the baseline dataset.}
    \label{fig:training_dataset_meters_distribution}
\end{figure}

Figure \ref{fig:training_dataset_meters_distribution} shows the meter distribution for the dataset. As can be noticed from the figure, the dataset contains \textit{verses} for each of the well-defined meters in \textit{Aroud} science. Each \textit{verse} accompanies its transcription. However, in Arabic poetry, some meters are more popular than others, especially the "\textit{Taweel}" meter. This bias is reflected in the meter distribution in the dataset.

\begin{table}[]
\centering{%
\begin{tabular}{|l|l|}
\hline
Number of annotated files       & 3,668   \\ \hline
Total size in hours             & 9.277  \\ \hline
Average files length in seconds & 8.985  \\ \hline
Total storage size              & 897 MB \\ \hline
\end{tabular}%
}
\caption {Some important statistics related to the baseline dataset.}
\label{table:training_dataset_statistics}
\end{table}

Regarding recording statistics, the dataset includes poems recited by multiple reciters. Each reciter recites a complete poem, generally. The reciters' information is not tracked in the dataset. Lab-settings details of the recording are also very limited. For example, we do not know who recites what and the total number of reciters. The reciters' gender information is also unavailable.

To train our models, we partition the dataset into almost 90\%, and 10\% for train and test splits. The test split here was not randomly chosen. Since speaker information is not available, we tried our best to selectively choose test samples recited by speakers not included in the training set. Although a separate test benchmark dataset is introduced in this research, reporting results on a blind portion of the same dataset could be valuable information to compare against the benchmark set. The resulting size after this partition is 3,309 and 359 \textit{Baits} for each split, respectively.

\subsection{The Benchmark Evaluation Set}

This dataset is a benchmark set containing 268 audio samples. Each sample is a \textit{Bait} that belongs to a meter. This benchmark is collected to evaluate the generalization of the proposed models. This dataset is publicly available. It can also be used as an evaluation benchmark for future works on this task.

To collect this dataset, we followed the following procedure: First, we skimmed the YouTube platform to select videos reciting Arabic poetry. This style of recitation is quite popular where the reciter recites the poem with a musical background. Usually, the recited poem text is provided in the video description. The audio is, then, extracted from the video. Since we are only interested in the vocal part, we remove the musical background, if any, and reduce any other background noise using the \citet{tsurumes91_online} toolkit. After that, we split the audio into segments of \textit{Baits} based on the reciter pauses as the reciter usually keeps a small pause in-between each \textit{Bait}. We filtered and collected good samples out of this process. We noticed that some meters do not have samples or only have very few samples. We decided to manually recite and upsample these unpopular meters.

\begin{figure}
    \centering
    \includegraphics[width=11cm,height=11cm,keepaspectratio]{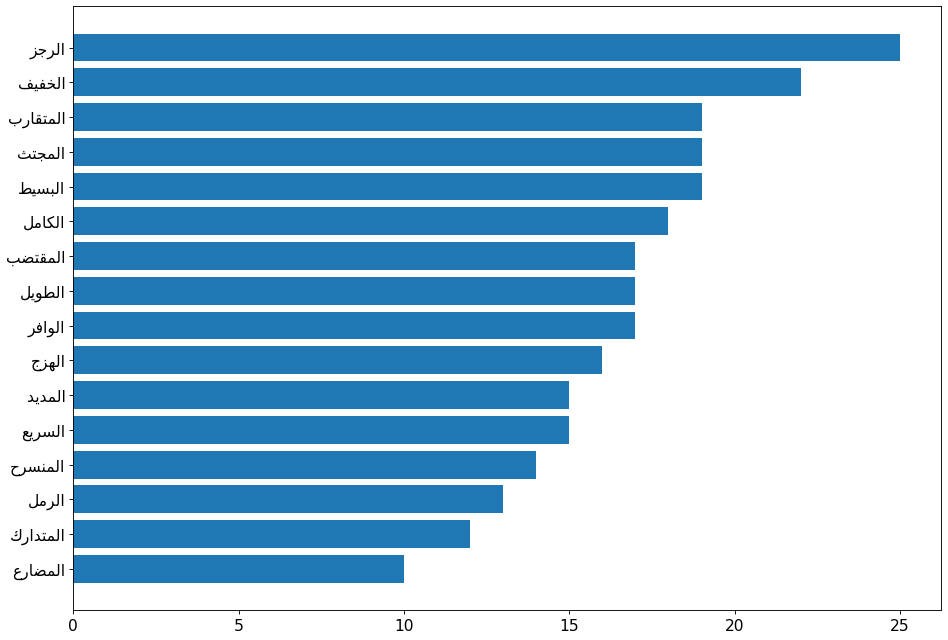}
    \caption{Meter distribution for the 16 poem meters in the benchmark set.}
    \label{fig:benchmark_dataset_meters_distribution}
\end{figure}

The meter distribution of this set is illustrated in Figure \ref{fig:benchmark_dataset_meters_distribution}. Further information and statistical details about this benchmark set are provided in Table \ref{table:benchmark_dataset_statistics}. As a criterion, while collecting this dataset, a minimum of 10 samples for each meter was collected. The maximum number of samples per meter should not be far from the minimum sample. This results in having a maximum number of 25 verses per meter. Although this benchmark, with this distribution, is balanced, its distribution is different from the training set distribution.

\begin{table}[]
\centering{%
\begin{tabular}{|l|l|}
\hline
Number of annotated files       & 268    \\ \hline
Total size in hours             & 0.48   \\ \hline
Average files length in seconds & 6.442  \\ \hline
Total storage size              & 105 MB \\ \hline
\end{tabular}%
}
\caption{Some important statistics related to the benchmark set.}
\label{table:benchmark_dataset_statistics}
\end{table}

\section{Methodology}
\label{methodology}

To investigate this problem, we tried two approaches, an end-to-end approach, and a transcription-based approach. Before discussing these architectures, we present an overview of the evaluation metrics we employed throughout this research. The next section introduces such metrics. The following sections present a detailed discussion of each of the proposed architectures.

\subsection{Evaluation Metrics}

Six metrics are reported in this research. Three of these metrics are classification-specific while the remaining two metrics are well-known in transcription and recognition tasks.

\subsubsection{Classification Metrics}

For a set of samples $S$ that belong to a set of classes $C$. A classification system gives predictions $P$ for these samples. The True Positives $TP$ samples of a class, let us name it $c$, are the labels predicted as $c$ by the system. Similarly, the True Negatives $TN$ are the non $c$ samples predicted as not $c$. False Positives $FP$ and False Negatives $FN$ are just the contrary. The measure the performance of this system we report the following metrics.

\begin{itemize}
    \item Accuracy ($A$): $$\frac{TP+TN}{TP+TN+FP+FN} \times 100$$
    \item Precision ($P$): $$\frac{TP}{TP+FP} \times 100$$
    \item Recall ($R$): $$\frac{TP}{TP+FN} \times 100$$
    \item F1--score: $$2\times \frac{P \times R}{P+R} \times 100$$ 
\end{itemize}

\subsubsection{Metrics to evaluate text transcription results}

A transcription or recognition system that recovers text from another source can be evaluated by many metrics \cite{morris2004and}. However, we will focus on the most well-known ones in this research: Word Error Rate (WER) and Character Error Rate (CER). WER calculates the percentage of words restored incorrectly. CER performs similar calculations but on the character's level. To compute these metrics, we used the implementation provided by \citet{jitsijiw8:online}.

\subsection{An End-to-End System for Poem-Meter Classification}
\label{sec:end2end}

We develop an end-to-end model that takes an audio sample as an input and produces the poem-meter as an output. A straightforward approach would be to train a system from scratch using labelled data. However, this was not feasible due to the lack of large datasets. Due to this limitation, we opted for transfer learning and fine-tuned a pretrained model.

We selected a pretrained Arabic speech recognition system to adapt it for poem-meter classification task. As suggested by \citet{yousef2019learning,al2020meter,alyafeai2022evaluating} and other works in this field, character sequences are the best features to classify Arabic poem meters. Based on this observation, we used Wav2Vec2 \citep{baevski2020wav2vec}, a state-of-the-art transformer-based \citep{vaswani2017attention} deep learning system for speech recognition at the character level. 

\begin{figure}
    \centering
    \includegraphics[width=12cm,height=14cm,keepaspectratio]{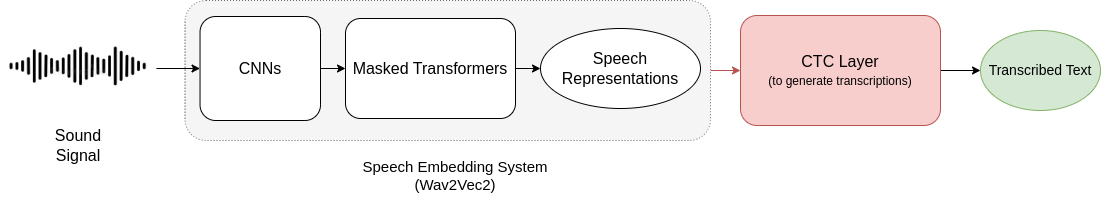}
    \caption{A high-level overview of Wav2Vec2 architecture.}
    \label{fig:Wav2Vec2}
\end{figure}

Wav2Vec2 is a transformer-based model that tries to learn speech representation from raw audio. The architecture consists of the following: first, the raw audio is fed into a CNN layer. The purpose of this layer is to down-sample the raw signal to be easily learnt by the transformer layer. This convolution operation produces latent space representation of the raw audio. A quantization operation is applied to this latent space. The quantized vectors are masked and fed to a transformer layer to predicted the masked vectors. The loss function used to train the architecture is the contrastive loss where it restricts the model to predict masked vectors that are close to the true ones and as far away as possible from other vectors. Figure \ref{fig:Wav2Vec2} overviews the architecture of the Wav2Vec2 system.

Wav2Vec2 model has been trained on speech from many languages including Arabic. There are many pretrained models proposed for Arabic. We selected the one that achieved the best performance in terms of Character Error Rate (CER) and Word Error Rate (WER) \cite{bakriano30_online}.

\begin{figure}
    \centering
    \includegraphics[width=12cm,height=14cm,keepaspectratio]{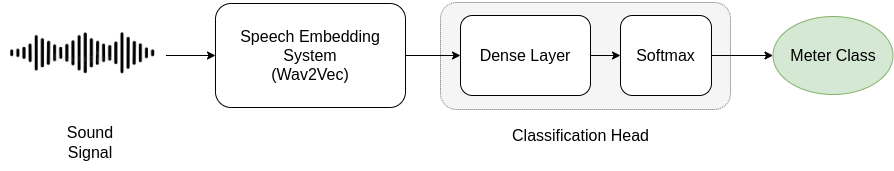}
    \caption{An illustration of the end-to-end system for poem-meter classification.}
    \label{fig:end_to_end_framework}
\end{figure}

We customized it to develop our end-to-end model for poem-meter classification. We removed the last CTC \citep{graves2006connectionist} layer from the model and extract only the character features, as the CTC layer is responsible for classifying these features into characters. We incorporated a classification head consisting of a dense layer followed by a softmax layer on top of this model to perform the meter classification. Figure \ref{fig:end_to_end_framework} illustrates the system. We evaluated this system by fine-tuning it using the baseline dataset and test it on the test sets of the two datasets introduced in Section \ref{dataset}. 

\subsection{Transcription-Based Poem-Meter Classification by Integrating Two Separate High-Resource Systems} 

As shown in Section \ref{sec:end2endResults}, the performance of the end-to-end system fails to achieve high accuracy. The main reason behind this is the lack of enough training data to robustly train the system.  
Since the dataset size we have is relatively small compared to the dataset sizes for the meter-classification systems trained on textual data (cf. \cite{al2020meter,abandah2020classifying}), it was expected that this approach may not produce the best results. In order to overcome this data scarcity issue, we propose an alternative framework that integrates two separate systems trained for two different high-resource tasks to perform poem-meter classification on recited poetry, which is a low-resource task.

We first employ a fine-tuned speech recognition system that transcribes the poem audio into text and then classify the transcribed text to poem meters using another system that is trained to classify poem meters from textual poems. We used the Wav2Vec2 \citep{baevski2020wav2vec} introduced in Section \ref{sec:end2end} for transcribing the recited poetry at the character level. The poetry text is then fed to another pretrained system for poem-meter classification. A number of Arabic poem meter classification systems exist in the literature that uses the textual poem verses as input, such as \citet{al2020meter}, \citet{yousef2019learning} and \citet{abandah2020classifying}. We choose the system presented in \cite{abandah2020classifying} because it includes all the 16 Arabic poem meters in addition to the prose class.

\subsubsection*{Speech Recognition Model Fine-Tuning}

As the Wav2Vec2 speech recognition system was not specifically trained for transcribing recited poems, we fine-tuned the model on the training set of the baseline dataset, randomly separating 10\% of it for validation. The motivation behind this is that Wav2Vec shows promising results when trained on a relatively small dataset compared to contemporary datasets such as the common voice dataset \citep{ardila2019common}, as shown by \citet{vaessen2022training}. 
For that purpose, we fine-tuned the Wav2Vec2 model for 10 epochs with 8 gradient-accumulation steps resulting in more than 1800 update steps. We used this fine-tuned model as a transcription model and compared the results with the original model. Throughout this research, we will refer to this model as Wav2Vec-FineTuned and the original model as Wav2Vec-Base.

\begin{figure}
    \centering
    \includegraphics[width=12cm,height=14cm,keepaspectratio]{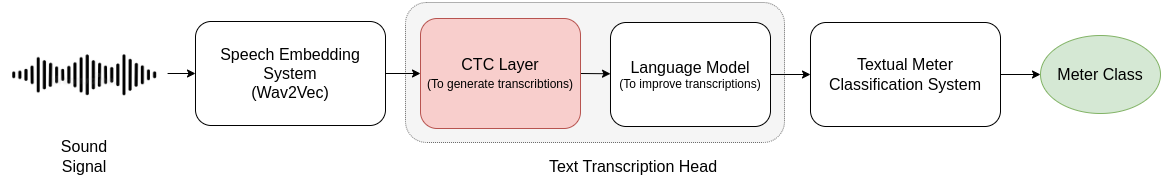}
    \caption{An integrated system for transcription-based poem meter classification.}
    \label{fig:transcription_based_framework}
\end{figure}

\subsubsection*{Language Model Head}

To further improve the quality of the transcribed text, we added a language model head on top of the speech recognition system to improve the transcription before feeding it into the text-based meter classification model. We employed a word 4-gram trained on poem texts. The choice of using a 4-gram language model was based on the observation that a poem verse in Arabic comprises two parts "\textit{Shatr}", with 4 words on average for each part. The language model was trained on the \citet{PCD2018} dataset as it is the largest available Arabic poetry dataset. We used the KenLM \citep{heafield2011kenlm} framework to build the language model in the conventional ARPA format.

\section{Results and Analysis}
\label{results}

In this section, we present the results of the experiments conducted in this research. We also provide an analysis and discussion on these results. We dedicate a subsection for each category of experiments.

\subsection{Results using the End-to-End System for Poem-Meter Classification}
\label{sec:end2endResults}

In this approach, a classification head is built on top of Wav2Vec2. This classification head is trained on the training set of the baseline dataset. Table \ref{tab:end_to_end_results_on_test_sets} summarizes the results achieved by this model on the test split of the baseline dataset. As can be noticed from the table, the model achieved 40\% accuracy on the test set of the baseline dataset, with the F1--score of 0.37. These results, although better than a random guess, are clearly very low.


\begin{table}[]
\centering
\begin{tabular}{ll}
\hline
Accuracy  & 40\% \\ \hline
Precision & 40\% \\ \hline
Recall    & 42\% \\ \hline
F1-Score  & 37\% \\ \hline
\end{tabular}
\caption{Results of end-to-end meter classification results on the baseline dataset (test split)}
\label{tab:end_to_end_results_on_test_sets}
\end{table}

\begin{figure}
    \centering
    \includegraphics[width=0.9\textwidth]{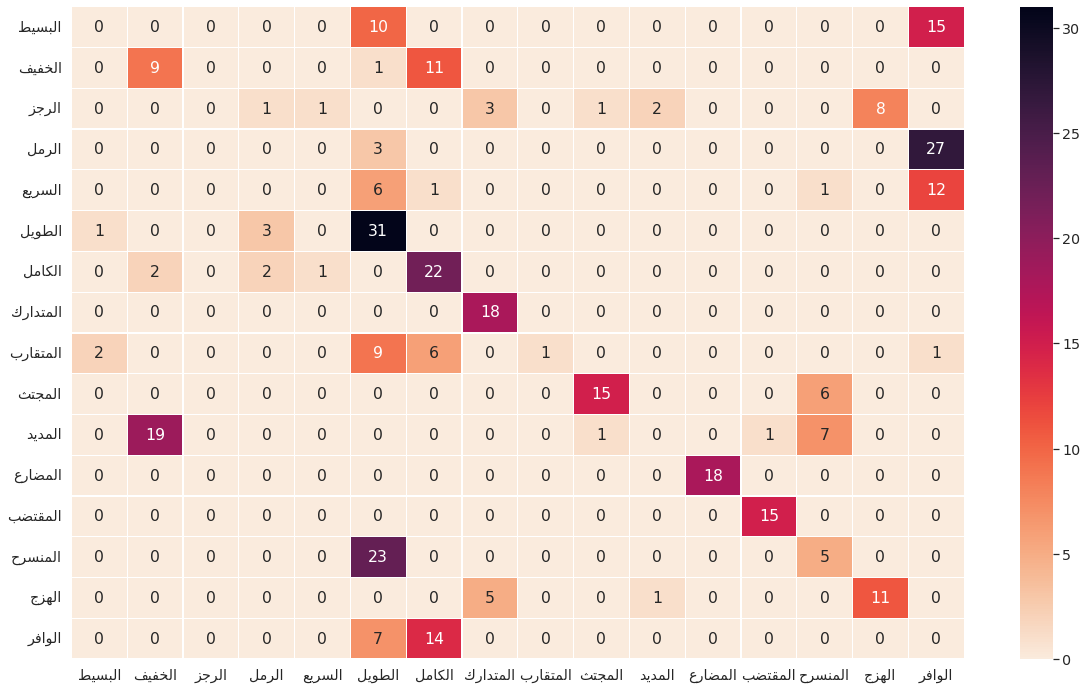}
    \caption{Confusion matrix for poem-meter classification using the End-To-End system on the baseline dataset.}
    \label{fig:end_to_end_benchmark_dataset_classification_heatmap}
\end{figure}

For further illustration and analysis of this experiment, we present the confusion matrix of the end-to-end model evaluated on the test split of the baseline dataset in figure \ref{fig:end_to_end_benchmark_dataset_classification_heatmap}. As can be seen from the matrix, some classes are classified with high accuracy. Many others, in contrast, got very poor classification results. We hypothesized that the model may learn other properties about the dataset rather than the meter, such as speakers' details. These results strongly suggest a larger dataset to fine-tune. To further verify the usefulness of the current dataset size for this task, we trained the architecture proposed by \citet{al2020meter} for textual meter classification on the ground-truth transcriptions we have in the training set of the baseline dataset and evaluate the results on the test set of it. The model achieved a low accuracy of 25\%. Hence, it is not surprising that this approach with the current dataset size does not provide promising results. However, with a much larger dataset, it may produce better results.

\subsection{Results Using the Transcription-Based Approach}


In this approach, as illustrated in figure \ref{fig:transcription_based_framework}, the signal is fed to Wav2Vec2 to generate the embeddings. These embeddings are then forwarded to the CTC layer to transcribe them to text. Aiming to improve the transcription, the transcribed text is forwarded to the language model layer to improve the transcription. The transcribed text is then fed to a meter classification model and the meter class is finally predicted for the recited verse.

To investigate the effect of the model fine-tuning and the use of the language model on the performance of poem transcription, we conducted a separate set of experiments using different system configurations, with and without the language model. In Table \ref{tab:transcription-classification-results}, we summarize the poem transcription results for both Wav2Vec-Base and Wav2Vec-FineTuned models in terms of Word Error Rate (WER) and Character Error Rate (CER) on both datasets. As can be seen from the table, the language model significantly improved the results on both the base as well as the fine-tuned models in terms of the WER and CER. 
The other important observation from the table is that fine-tuning with language models slightly improved the results on the Baseline dataset but degrades the result on the benchmark set. This observation can be explained by the fact that the Wav2Vec model was fine-tuned on the training set of the Baseline dataset. The Benchmark set, which was scraped from the web, seems to have different properties with regard to audio quality as compared to the Baseline dataset, which was collected within a lab setting. Thus, fine-tuning the general-purpose speech recognizer on the Baseline dataset seems to adversely affect the performance on the Benchmark set.


\begin{table}[]
\centering
\begin{tabular}{lllll}
\multicolumn{1}{c}{\multirow{2}{*}{}}          & \multicolumn{2}{l}{Baseline (test set)} & \multicolumn{2}{l}{Benchmark set} \\ \cline{2-5} 
\multicolumn{1}{c}{}                           & WER     & \multicolumn{1}{l|}{CER}     & WER            & CER          \\ \hline
\multicolumn{1}{l|}{Wav2Vec2Base}              & 21.8\%  & \multicolumn{1}{l|}{4.8\%}  & 24.9\%         & 7\%          \\ \hline
\multicolumn{1}{l|}{Wav2Vec2Base with LM}      & 11.1\%  & \multicolumn{1}{l|}{\textbf{2.6}\%}    & \textbf{14.7\%}           & \textbf{5\%}          \\ \hline
\multicolumn{1}{l|}{Wav2Vec2FineTuned}         & 22.7\%  & \multicolumn{1}{l|}{5.3\%}  & 36.2\%         & 10.4\%         \\ \hline
\multicolumn{1}{l|}{Wav2Vec2FineTuned with LM} & \textbf{10.7\%}  & \multicolumn{1}{l|}{2.7\%}  & 22\%           & 7.8\%        \\ \hline
\end{tabular}
\caption{Poem-transcription results using different system configurations on the two datasets.}
\label{tab:transcription-classification-results}
\end{table}


We next move to the poem-meter classification results where we use the transcription output of the first system and use that to predict the poem meter by feeding it to a system for poem-meter classification on textual poems. In 
Table \ref{tab:transcription_based_framework_results_wav2vec_base}, we summarize the results of the Wav2Vec-Base model integrated with the textual poem-meter classification system. 
The behavior of the language model on both test sets is also reported. As can be seen in the table, the results on the Baseline dataset are lower than the Benchmark set. These results suggest enlarging the dataset with the same lab settings for better results.


The behavior of the language model on both test sets is also reported. From the table, a consistent margin of improvement due to the use of the language model can be noticed on both sets across all measures. For instance, the accuracy was improved with LM by a margin of approximately 2\% for both sets. 

\begin{table}[]
\centering
\begin{tabular}{|c|c|cccc|}
\hline
\multirow{2}{*}{Dataset}                                                 & \multirow{2}{*}{}           & \multicolumn{4}{c|}{Wav2Vec-Base}                                                                                             \\ \cline{3-6} 
                                                                         &                             & \multicolumn{1}{c|}{Accuracy}      & \multicolumn{1}{c|}{Precision}     & \multicolumn{1}{c|}{Recall}        & F1-Score      \\ \hline
\begin{tabular}[c]{@{}c@{}}Baseline dataset\\ (Test set)\end{tabular} & \multirow{2}{*}{Without LM} & \multicolumn{1}{c|}{85\%}          & \multicolumn{1}{c|}{83\%}          & \multicolumn{1}{c|}{80\%}          & 79\%          \\ \cline{1-1} \cline{3-6} 
Benchmark set                                                            &                             & \multicolumn{1}{c|}{90\%}          & \multicolumn{1}{c|}{89\%}          & \multicolumn{1}{c|}{83\%}          & 85\%          \\ \hline
\begin{tabular}[c]{@{}c@{}}Baseline dataset\\  (Test set)\end{tabular} & \multirow{2}{*}{With LM}    & \multicolumn{1}{c|}{87\%}          & \multicolumn{1}{c|}{84\%}          & \multicolumn{1}{c|}{81\%}          & 80\%          \\ \cline{1-1} \cline{3-6} 
Benchmark set                                                            &                             & \multicolumn{1}{c|}{92\%} & \multicolumn{1}{c|}{90\%} & \multicolumn{1}{c|}{86\%} & 87\% \\ \hline
\end{tabular}
\caption{Meter classification results integrating the \textbf{Wav2Vec-Base} system for poem transcription and poem-text-to-meter classification system from \citet{abandah2020classifying}.}
\label{tab:transcription_based_framework_results_wav2vec_base}
\end{table}

Table \ref{tab:transcription_based_framework_results_wav2vec_finetuned} shows the results of the Wav2Vec-FineTuned model on both datasets. It can be seen that the fine-tuned model does not outperform the base model. This can be attributed to our training set distribution shift from the Wav2Vec2 pretrained training set. In order to improve the results, a larger dataset drawn from this distribution should be used.

\begin{table}[]
\centering
\begin{tabular}{|c|c|cccc|}
\hline
\multirow{2}{*}{Dataset}                                                 & \multirow{2}{*}{}           & \multicolumn{4}{c|}{Wav2Vec-FineTuned}                                                                   \\ \cline{3-6} 
                                                                         &                             & \multicolumn{1}{c|}{Accuracy} & \multicolumn{1}{c|}{Precision} & \multicolumn{1}{c|}{Recall} & F1-Score \\ \hline
\begin{tabular}[c]{@{}c@{}}Baseline dataset\\  (Test set)\end{tabular} & \multirow{2}{*}{Without LM} & \multicolumn{1}{c|}{85\%}     & \multicolumn{1}{c|}{83\%}      & \multicolumn{1}{c|}{80\%}   & 79\%     \\ \cline{1-1} \cline{3-6} 
Benchmark set                                                            &                             & \multicolumn{1}{c|}{82\%}     & \multicolumn{1}{c|}{82\%}      & \multicolumn{1}{c|}{75\%}   & 76\%     \\ \hline
\begin{tabular}[c]{@{}c@{}}Baseline dataset\\  (Test set)\end{tabular} & \multirow{2}{*}{With LM}    & \multicolumn{1}{c|}{86\%}     & \multicolumn{1}{c|}{84\%}      & \multicolumn{1}{c|}{81\%}   & 80\%     \\ \cline{1-1} \cline{3-6} 
Benchmark set                                                            &                             & \multicolumn{1}{c|}{88\%}     & \multicolumn{1}{c|}{86\%}      & \multicolumn{1}{c|}{82\%}   & 83\%     \\ \hline
\end{tabular}
\caption{Meter classification results integrating the \textbf{Wav2Vec-FineTuned} system for poem transcription and poem-text-to-meter classification system from \citet{abandah2020classifying}.}
\label{tab:transcription_based_framework_results_wav2vec_finetuned}
\end{table}

As in the Wav2Vec-Base experiment, the language model improved the transcription results. However, the improvement is much higher than Wav2Vec-Base on all measures for the benchmark set. For instance, the accuracy improvement was only 2\% for Wav2Vec-Base but it was improved by 6\% on Wav2Vec-FineTuned for the Benchmark dataset. This also follows for F1-Score and recall. These results stress the critical importance of the developed language model for this task with such settings.

\begin{figure}
    \centering
    \includegraphics[width=0.9\textwidth]{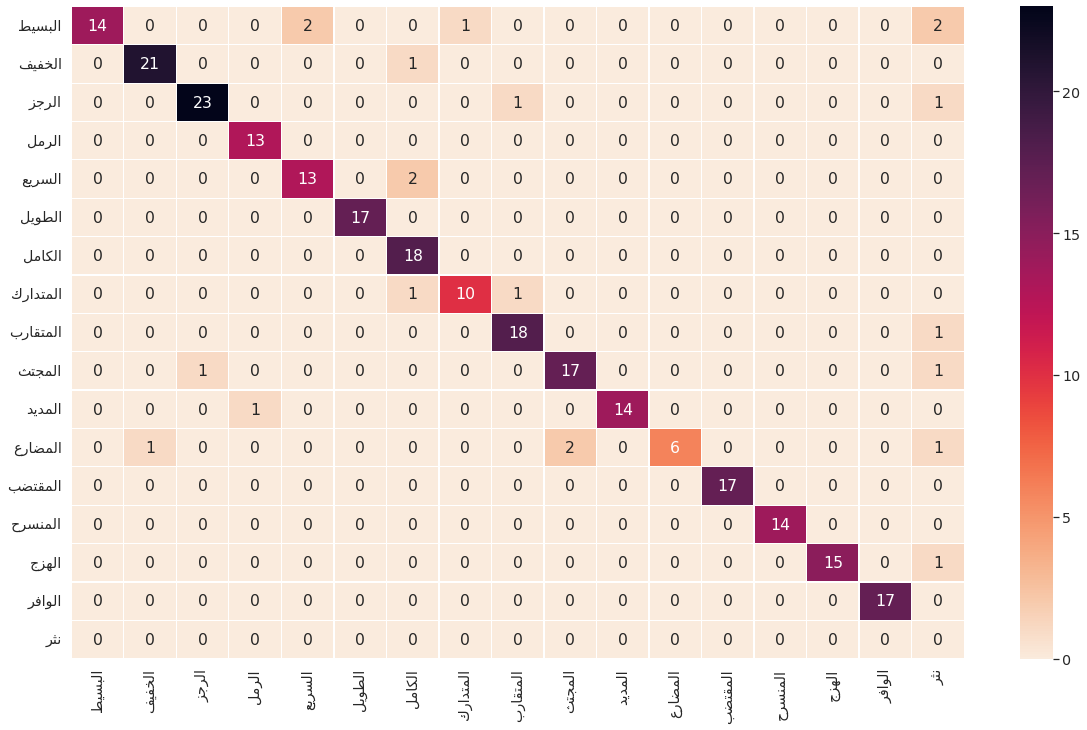}
    \caption{Confusion matrix for poem-meter classification using the \textbf{Wav2Vec-Base} system for poem transcription and poem-text-to-meter classification system from \citet{abandah2020classifying}.}
    \label{fig:transcription_based_with_lm_benchmark_dataset_classifcation_heatmap}
\end{figure}

\begin{figure}
    \centering
    \includegraphics[width=0.9\textwidth]{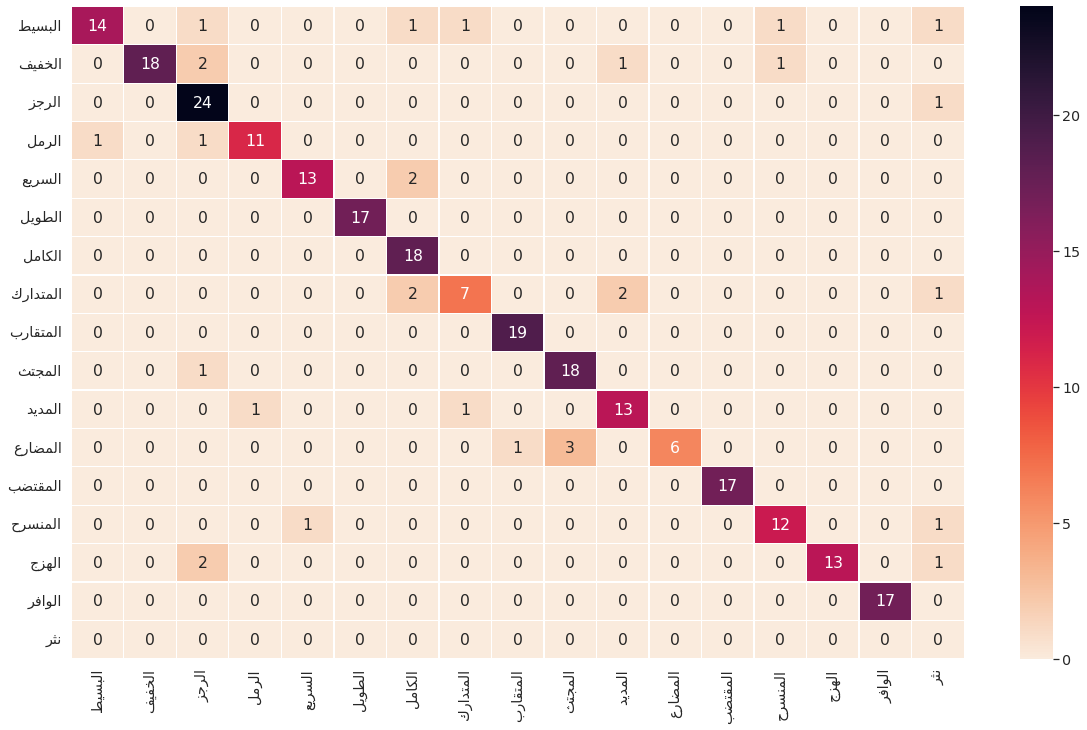}
    \caption{Confusion matrix for poem-meter classification using the \textbf{Wav2Vec-FineTuned} system for poem transcription and poem-text-to-meter classification system from \citet{abandah2020classifying}.}
    \label{fig:finetuned_transcription_based_with_lm_benchmark_dataset_classifcation_heatmap}
\end{figure}

For a deeper analysis of the model behaviors on the benchmark set, Figure \ref{fig:transcription_based_with_lm_benchmark_dataset_classifcation_heatmap} and Figure \ref{fig:finetuned_transcription_based_with_lm_benchmark_dataset_classifcation_heatmap} show the confusion matrices for the Wav2Vec-Base and Wav2Vec-Finetuned results on the benchmark set with the Language Model head, respectively. We chose the models with the language model because they show better results, as can be seen from Table \ref{tab:transcription_based_framework_results_wav2vec_base} and Table \ref{tab:transcription_based_framework_results_wav2vec_finetuned}. We also chose the benchmark set, as we are proposing this set as a standard benchmark for any further evaluation of this task.

As can be noticed from Figure \ref{fig:transcription_based_with_lm_benchmark_dataset_classifcation_heatmap} and Figure \ref{fig:finetuned_transcription_based_with_lm_benchmark_dataset_classifcation_heatmap}, Wav2Vec-Base made more errors classifying verses as prose than Wav2Vec-FineTuned. This indicates that Wav2Vec-FineTuned learns more about poetry than Wav2VecBsae as a result of the fine-tuning process. However, this observation can be further studied and validated when fine-tuning occurs with a larger baseline dataset. Based on our analysis, we can hypothesize that the fine-tuned model would be less confused with prose as compared to the Wav2Vec-Base model when trained on a larger dataset. 

Another observation from the reported confusion matrix is that both models are easily confused between close yet rare classes such as \textit{Mudhare} and \textit{Mujtath} (\RL{المضارع}, \RL{المجتث}). This observation is expected, as having fewer samples of a class would most probably confuse it with its closest class.

Another observation from both the confusion matrices is that the samples of the most represented classes in the training set, as depicted in figure \ref{fig:training_dataset_meters_distribution}, are almost correctly classified. For instance, \RL{الطويل}, \RL{الرمل}, \RL{الوافر}, \RL{الكامل}, and  \RL{المنسرح} were classified correctly with Wav2Vec-Base. The same behavior is noticed with Wav2Vec-FineTuned with a minor error margin. This can be deduced as a result of two main reasons: these classes are also widely used by poets in Arabic poetry, and there is a high possibility that many samples belonging to these classes are in the Wav2Vec baseline dataset. The second reason may only apply for \RL{المنسرح} which has a rhythmic structure that is distinct from other classes.

\subsubsection{Ablation study}

In this section, we report the results of an ablation study we carried out to assess the impact of each sub-system on meter-classification performances. As mentioned before, we integrate two separate high-resource systems for the task of poem-meter classification of the recited poems. A portion of the errors is due to the mistakes from poem transcription system. Text-based poem-meter classification system accounts for the remaining error. The poem transcription system's performance in terms of WER and CER was presented in Table \ref{tab:transcription-classification-results}. However, this does not give a complete picture on how much of those errors lead to mistakes in poem-meter classification. Thus, in the current ablation study we provided the perfect annotation for each audio sample by directly using the ground-truth information and feeding them to the textual poem-meter classification system. The results on the two datasets are presented in Table \ref{tab:ground_truth_results}. From the table we can see that the system achieves F--1 score of 87\% and 89\% on the Baseline and Benchmark sets, respectively. Our best system achieves an F--1 score of 80\% on the Baseline dataset and  87\% on the Benchmark dataset (cf. Tables \ref{tab:transcription_based_framework_results_wav2vec_base} and \ref{tab:transcription_based_framework_results_wav2vec_finetuned}). Thus, the poem transcription system accounts for almost 7\% drop in F--1 score on the Baseline set and the remaining 13\% drop is from the meter classification system. On the Bechmark set, only 2\% drop in F--1 score is due to the poem transcription system and around 11\% drop is due to the meter classification system.

\begin{table}[]
\centering
\begin{tabular}{lcccc}
\hline
\textbf{Dataset} & \textbf{Accuracy} & \textbf{Precision} & \textbf{Recall} & \textbf{F1--Score} \\ \hline
Baseline (test split) & 89\%     & 91\%      & 88\%   & 87\%      \\ \hline
Benchamrk             & 95\%     & 91\%      & 88\%   & 89\%      \\ \hline
\end{tabular}
\caption{Poem-meter classification results using perfect speech transcriptions.}
\label{tab:ground_truth_results}
\end{table}

In summary, the end-to-end approach did not provide promising results given the size limitation of the current dataset. However, the transcription-based approach provides state-of-the-art results on this task. The transcriptions provided by the Wav2Vec2 architecture can be considered state-of-the-art for this task. It is, nevertheless, recommended to be improved via a language model. The proposed models for textual meter classification are reliable enough to circumvent transcription mistakes. With WERs above 10\% and CERs between 2--5\%, the poem-meters can still be robustly identified. 

\section{Conclusions and future work}
\label{conclusion}

In this work, two systems to classify Arabic poetry from acoustic sources are introduced. Both of these systems are based on Wav2Vec2, a transformer-based speech representation. The first system is an end-to-end system, while the second system is a transcription-based system that integrates two high-resource systems to perform the low-resource task of poem-meter classification of recited Arabic poetry. We also introduced a public benchmark evaluation set to benchmark future research on this task. The end-to-end system did not perform very well mainly due to the lack of enough training data. The idea of integrating two separate high-resource systems to perform a low-resource task seemed to work in this case. The system was evaluated to two separate datasets with F1--scores ranging between 80\% and 87\%.

As a future work, following the conclusions drawn in this research, a larger dataset of recited Arabic poems should be introduced for this task. Furthermore, it will be interesting to study the performance of other state-of-the-art speech recognition models, such as the Whisper model \citep{radford2022robust}, for transcription-based tasks. One may also investigate techniques to normalize sound signals to minimize the effects of audio-processing software. Last but not the least, data augmentation techniques can be explored to reduce the model overfitting problem and resolving data imbalance, at the same time. 

\section*{Data Statement}
The datasets generated during and/or analysed during the current study are available from the corresponding author on reasonable request. The baseline dataset used in this research is a private dataset. The dataset could be provided for research purposes. However, it cannot be used to produce models for commercial use.

\section*{Conflict of Interest}
None.

\section*{Acknowledgment}
The authors would like to thank King Fahd University of Petroleum \& Minerals (KFUPM) for supporting this work. Irfan Ahmad is supported by SDAIA-KFUPM Joint Research Center for Artificial Intelligence through grant number JRC--AI--RFP--06.

\bibliographystyle{sn-basic}
\bibliography{refs}

\end{document}